\documentclass[letterpaper, 10 pt, conference]{ieeeconf}  

\IEEEoverridecommandlockouts                              

\usepackage{graphicx}
\usepackage{amsmath}
\usepackage{amssymb}
\usepackage{booktabs}
\usepackage{float}
\usepackage{color}
\usepackage{subcaption}

\title{\LARGE \bf
Revisiting Event-based Video Frame Interpolation
}

\author{Jiaben Chen$^{1}$, Yichen Zhu$^{2}$, Dongze Lian$^{3}$, Jiaqi Yang$^{2}$, Yifu Wang$^{2}$, Renrui Zhang$^{4}$, Xinhang Liu$^{2}$, \\ Shenhan Qian$^{5}$, Laurent Kneip$^{2}$ and Shenghua Gao$^{2}$
\thanks{This work was supported by NSFC \#61932020, \#62172279, \#62250610225, Program of Shanghai Academic Research Leader, projects 22dz1201900 and 22ZR1441300 funded by the Shanghai Science Foundation and “Shuguang Program” supported by Shanghai Education Development Foundation and Shanghai Municipal Education Commission.}
\thanks{$^{1}$UC San Diego; $^{2}$ShanghaiTech University; $^{3}$National University of Singapore; $^{4}$Shanghai AI Laboratory; $^{5}$Technical University of Munich}
}

\begin{document}

\maketitle
\thispagestyle{empty}
\pagestyle{empty}

\begin{abstract}

Dynamic vision sensors or event cameras provide rich complementary information for video frame interpolation. Existing state-of-the-art methods follow the paradigm of combining both synthesis-based and warping networks. However, few of those methods fully respect the intrinsic characteristics of events streams. Given that event cameras only encode intensity changes and polarity rather than color intensities, estimating optical flow from events is arguably more difficult than from RGB information. We therefore propose to incorporate RGB information in an event-guided optical flow refinement strategy. Moreover, in light of the quasi-continuous nature of the time signals provided by event cameras, we propose a divide-and-conquer strategy in which event-based intermediate frame synthesis happens incrementally in multiple simplified stages rather than in a single, long stage. Extensive experiments on both synthetic and real-world datasets show that these modifications lead to more reliable and realistic intermediate frame results than previous video frame interpolation methods. Our findings underline that a careful consideration of event characteristics such as high temporal density and elevated noise benefits interpolation accuracy.

\end{abstract}

\section{INTRODUCTION}

Video Frame Interpolation (VFI), the task of converting low frame rate videos into high frame rate sequences by interpolating new frames between two given consecutive frames \cite{niklaus2017video,niklaus2018context}, has promising applications in the field of robotics. By providing additional visual information, it can enhance robot motion planning and control. Although dedicated hardware for capturing high frame-rate videos exists, it would be costly to deploy such devices.

\begin{figure}[h!]
\setlength{\belowcaptionskip}{-0.3cm}
\begin{center}
\includegraphics[width=\linewidth]{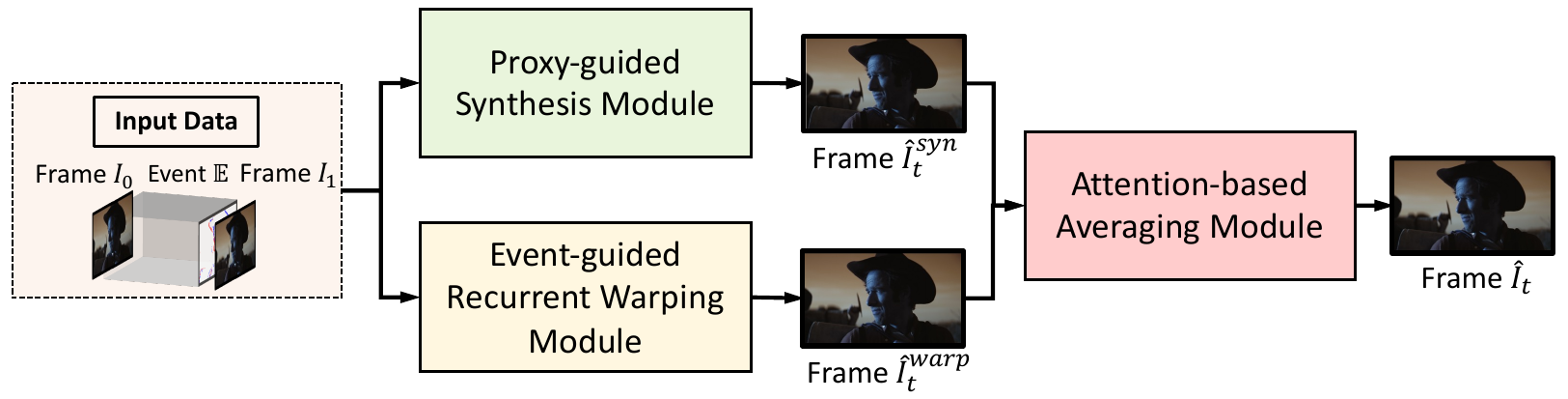}
\end{center}
\caption{Overview of the proposed architecture for event-supported video frame interpolation. It employs the state-of-the-art concept of fusing results from warping and synthesising networks, and carefully respects the characteristics of events in the internal network design.}
\label{fig:overview}
\end{figure}
In RGB image-based VFI, the leveraged information remains limited by the frame rate of the original videos. Recently, event cameras have proven their ability to complement RGB imagery with temporally dense sensing, thus providing additional constraints to improve traditional VFI \cite{lin2020learning,han2021evintsr,yu2021training}. This paper aims at event-supported VFI. 

Existing work such as Time Lens \cite{tulyakov2021time} already utilizes an event camera for event-based VFI. It achieves remarkable performance by combining a synthesis and a warping network in a two-stream architecture. The synthesis network directly synthesizes intermediate frames, while the warping network utilizes events to generate optical flow for frame warping. However, Time Lens \cite{tulyakov2021time} solely employs events to estimate optical flow in the warping network. We argue that directly estimating optical flow from events is a very challenging task, and inferior results can compromise the performance of VFI. Furthermore, Time Lens \cite{tulyakov2021time} utilizes only the outer image timestamps within the synthesis network and does not consider the fact that the event camera captures the image change in a continuous, high-frequent way. Consequently,  the full potential of including events has not yet been unlocked. To better leverage event traits within VFI, we propose the following two improvements.

First, event cameras only encode changes of intensity and polarity rather than absolute color intensities \cite{gallego2020event}. Estimating optical flow from events only is therefore arguably more difficult than from RGB cameras. RGB-based optical flow methods evaluate pixel value similarities between consecutive frames. In contrast, in event sequences, corresponding pixels present similar numbers of accumulated events within relatively small time intervals. The high sensitivity of event cameras leads to an abundance of events within short time intervals and non-negligible noise accumulations. On the other hand, spatial event distribution is often sparse and lacking clear information in low-textured regions. The strength of events is therefore considered to lie in their ability to sense change that occurs over very short periods of time. Over longer time periods, however, the accumulation of noise will blur information and complicate the identification of accurate correspondences. The calculation of optical flow solely from event sequences therefore appears to be more challenging than traditional image-matching-based approaches.

We argue that \textbf{event sequences are utilized better to calibrate optical flow estimated from RGB images rather than to estimate optical flow solely from an event camera}. To this end, we propose an \emph{event-guided recurrent warping} strategy, which more closely respects the intrinsic properties and advantages of event cameras.

Second, different from RGB cameras that capture intensities at a low frame rate, event cameras \cite{lichtsteiner2008128} record per-pixel intensity changes asynchronously at high, quasi-continuous temporal resolution. In the synthesis network, Time Lens \cite{tulyakov2021time} directly synthesizes the middle frame in one step. The rich continuous-time nature of events is not fully utilized thus leading to sub-optimal interpolation results. Motivated by global and local temporal feature fusion strategies in other video processing problems \cite{feichtenhofer2019slowfast}, we propose to replace the one-step method to obtain a single intermediate frame by a \emph{proxy-guided synthesis} strategy that synthesizes multiple intermediate frames in a step-by-step manner. Specifically, by slicing an event sequence along time, we can synthesize intermediate frames (\emph{proxies}) from thinner event segments. 
Our main rationale behind this design is \textit{divide and conquer}. \textbf{By splitting the events into multiple slices and synthesizing in stages, each stage is demanded to regress from a simpler function with reduced non-linearity using data that is less affected by temporal error accumulations}. Rather than solving a large difficult problem over a global time interval and implicitly understanding complicated non-linear motion over time, we propose to solve the problem sequentially over multiple local intervals containing simpler motion. The proposed \emph{proxy-guided synthesis} intuitively makes better use of the rich continuous-time nature of events. 

The overall architecture of our proposed framework for event-based VFI is shown in Fig. \ref{fig:overview}, and our main contributions are summarized as follows:
\begin{itemize}
    \item We propose an \emph{Event-guided Recurrent Warping} strategy, which uses events to calibrate optical flow estimates from RGB images. Our solution better grasps the advantages of event sequences.
    \item Exploiting the rich quasi-continuous nature of the temporal information provided by events, we design a \emph{Proxy-guided Synthesis} strategy to incrementally synthesize intermediate frames and effectively combine both local and global temporal information. 
    \item Extensive experiments on both synthetic and real-world benchmarks demonstrate VFI performance improvement by our approach.
\end{itemize}

\section{Related Work}
\subsection{Frame-based Interpolation}
Most existing frame-based VFI methods estimate intermediate frames with sparse input RGB frames at a fixed low rate, categorized into kernel-based and flow-based methods.

\noindent \textbf{Kernel-based approaches} model VFI as a local convolution of input key frames by predicting a spatially-adaptive convolutional kernel instead of estimating explicit intermediate motion \cite{niklaus2017video}. However, CNN-based fail at handling large displacements due to limited kernel size, or otherwise induce exploding computation times and memory consumption. \\
\noindent \textbf{Flow-based approaches} proceed by estimating optical flow between input frames and generate intermediate frames using image warping \cite{jaderberg2015spatial}. Linear \cite{jiang2018super,li2020video}, quadratic \cite{xu2019quadratic} and cubic \cite{chi2020all} trajectory assumptions have been made to approximate intermediate motion. Techniques like softmax splatting \cite{niklaus2020softmax}, motion fields \cite{park2020bmbc,park2021asymmetric}, Transformers \cite{lu2022video}, privileged distillation \cite{huang2022real}, and extra information like contextual maps \cite{niklaus2018context}, temporal sliding windows of frames \cite{kalluri2020flavr} and depth maps \cite{bao2019depth} have all been adopted to improve flow estimation and interpolation accuracy. Although these methods present great variety in their designs and approaches to handle complex motion, they are all limited by low-order assumptions within their inner flow estimator, which can be attributed to the lack of intermediate motion signals.

In a nutshell, frame-based approaches rely on the brightness constancy assumption and use limited visual information, thus leading to limited performance in scenarios with complex non-linear motion and illumination changes.
\subsection{Event-based interpolation}
In contrast to conventional cameras, event cameras \cite{lichtsteiner2008128} are relatively new sensors that capture temporally dense motion signals as a sequence of asynchronous per-pixel brightness changes. Owing to the rapid development of industrial event cameras, recent years have witnessed a surge of event-based methods addressing the VFI problem. Purely event-based methods learn video reconstruction from data and train deep neural networks like CNNs \cite{paredes2021back} and RNNs \cite{rebecq2019events}. More recently, the community has started to investigate the highly interesting sensor fusion variant of complementary event-plus-frame methods to solve the VFI problem\cite{wang2020event,pan2019bringing,lin2020learning,han2021evintsr,zhang2022unifying}. By capturing dense motion information and compressed true visual signals \cite{tulyakov2021time}, event cameras can serve as an ideal supplement for frame-based VFI in challenging scenarios. Following work \cite{tulyakov2022time} further explores using events to estimate non-linear motion between frames with a motion spline estimator. \cite{wu2022video} proposes to use events to blend optical flow from VFI flow estimator \cite{huang2022real}. Works like \cite{yu2021training,he2022timereplayer} explore potential of event-based VFI in weakly and unsupervised learning setting, respectively. \cite{tulyakov2021time} introduces a pipeline combining synthesis and warping-based approaches, showing promising results in high dynamic settings. However, their synthesis-based methods feed the whole event sequence directly into a synthesis network, thus leading to a loss of the rich temporal information provided by events and blurry intermediate frames. Furthermore, the method estimates optical flow solely from events which yields messy predictions when events are noisy, sparse or not correctly registered.






\section{Method}

\subsection{Preliminaries}

\noindent \textbf{Problem Definition.} 
Given consecutive video frames $I_0, I_1\in\mathbb{R}^{W \times H \times 3}$ as inputs, where $W$ and $H$ are the width and height of the frames, respectively, VFI aims to predict an intermediate new frame $\hat{I}_t$ at time $t \in (0, 1)$. For event-based VFI, event sequences $\mathbb{E}_{0 \rightarrow t}$ and $\mathbb{E}_{t \rightarrow 1}$ are also included as inputs, which encompass all triggered events between the time of the input RGB frame $I_0$ and the time $t$ of the intermediate frame, and the events between the latter and the input RGB frame $I_1$.

\noindent \textbf{Event Representation.} 
The events triggered in a given time interval form a sequence $\{e_i = (x_i, y_i, t_i, p_i)\}_{i \in [1,M]}$, where $M$ indicates the total number of events. Each event is comprised of a pixel position $x_i, y_i$ indicating the location of a brightness change, a timestamp $t_i$ indicating when the change occured, and a polarity $p_i$ indicating whether the perceived logarithmic brightness at that pixel increased or decreased by a certain threshold amount. Owing to their unique asynchronous nature, raw events cannot be directly taken as an input to a neural network. Following \cite{zhu2019unsupervised}, we adopt a discretized event volume representation by dividing the temporal dimension into $B$ bins. In this way, event sequences are encoded as a $B$-channel tensor permitting the application of 2D convolutions in spatial dimensions. All modules in our work use converted, voxel grid-based event representations.


\subsection{Overview of the architecture}
Drawing inspiration from Time Lens \cite{tulyakov2021time}, we propose a VFI framework that relies on two complementary modules: a synthesis module and a warping module. The overall architecture is shown in Fig. \ref{fig:overview}. It is mainly composed of three components: a \emph{Proxy-guided Synthesis} Module, an \emph{Event-guided Recurrent Warping} Module, and an \emph{Attention-based Averaging} Module. In the \emph{Proxy-guided Synthesis} Module, we slice the two event voxel grids between the boundary RGB frames ($I_0$ and $I_1$) and the desired intermediate frame $\hat{I}_t$ into two sub-slices along the temporal dimension, and we utilize these slices to incrementally synthesize further intermediate frames $\hat{I}^{\text{syn}}_t$, i.e., proxies (Sec. \ref{3.3}). In the \emph{Event-guided Recurrent Warping} Module, an optical flow estimate is first predicted from RGB frames and then refined by the event sequences. The output frame $\hat{I}^{\text{warp}}_t$ is generated by backward warping (Sec. \ref{3.4}). Finally, we output the interpolated frame $\hat{I}_t$ via an \emph{Attention-based Averaging} Module that blends the results of the previous two modules, thus aiming at overcoming individual deficiencies and combining the advantages of the synthesis-based and warping-based schemes (Sec. \ref{3.5}). A detailed exposition of each module is presented in the following sections.

\begin{figure}[b]
\begin{center}
    \includegraphics[width=\linewidth]{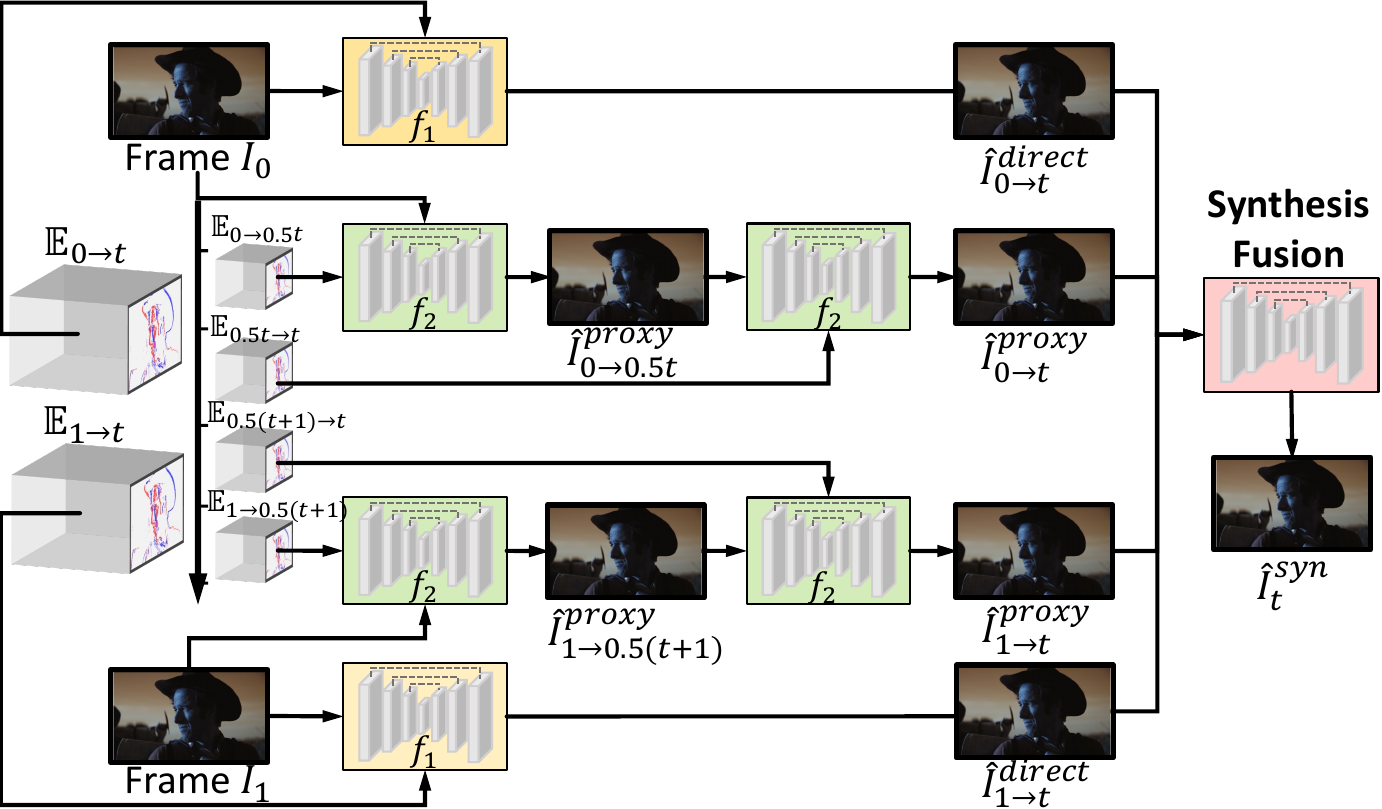}
\end{center}
\caption{Architecture of the \emph{Proxy-guided Synthesis Module}. The direct predictions of the middle frame are complemented by incremental predictions (\textbf{i.e.}, proxies) using fewer events obtained from smaller time intervals.} 
\label{fig:synthesis}
\end{figure}
\subsection{Proxy-guided Synthesis Module \label{3.3}}
\begin{figure}[H]
\setlength{\belowcaptionskip}{-0.4cm}
\begin{center}
\includegraphics[width=0.75\linewidth]{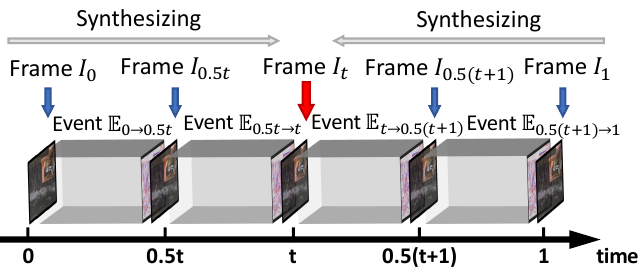}
\end{center}
\caption{Event segmentation over time within the \emph{Proxy-guided Synthesis Module}. The sequence is divided into subsequences to incrementally generate three intermediate frames.}
\label{fig:Segment}
\end{figure}

Aiming to take advantage of the rich continuous time signals provided by event cameras, we divide the event sequences along the time axis to obtain multiple event segments. More specifically, we split the synthesis module into two branches: the \emph{Direct Synthesis Branch}, and the \emph{Transitional Synthesis Branch}. The complete structure is shown in Fig. \ref{fig:synthesis}. The event segmentation strategy is illustrated in Fig. \ref{fig:Segment}.
\noindent \textbf{Direct Synthesis.} In the \emph{Direct Synthesis Branch}, we directly regress the intermediate frame from input RGB frames $I_0$ and $I_1$ as well as the two event sequences $\mathbb{E}_{0 \rightarrow t}$ and $\mathbb{E}_{1 \rightarrow t}$. Note that we utilize the event sequence reversal strategy from \cite{tulyakov2021time} to compute $\mathbb{E}_{1 \rightarrow t}$ from input sequence $\mathbb{E}_{t \rightarrow 1}$.

Taking $I_0$ and $\mathbb{E}_{0 \rightarrow t}$ as inputs, a neural network is employed to directly synthesise the desired intermediate frame $\hat{I}_{0 \rightarrow t}^{\text{direct}}$. Meanwhile, the intermediate frame is also generated in reverse direction by considering $I_1$ and $\mathbb{E}_{1 \rightarrow t}$ as inputs, denoted $\hat{I}_{1 \rightarrow t}^{\text{direct}}$:
\begin{equation}
\setlength{\abovedisplayskip}{5pt}
\setlength{\belowdisplayskip}{5pt}
    \left\{
    \begin{array}{cc}
         \hat{I}_{0 \rightarrow t}^{\text{direct}} &= f_1(I_0, \mathbb{E}_{0 \rightarrow t}) \\
         \hat{I}_{1 \rightarrow t}^{\text{direct}} &= f_1(I_1, \mathbb{E}_{1 \rightarrow t}),
    \end{array}
    \right.
\end{equation}
where $f_1$ is a neural network function. The \emph{Direct Synthesis Branches} generate intermediate frames using global time signals, which is insufficient to fully exploit the rich continuous-time nature of the events. To enhance the interpolation, we further consider more local time signals by our \emph{Transitional Synthesis Branch}.  

\noindent \textbf{Transitional Synthesis.} In the \emph{Transitional Synthesis Branch}, we learn from local time intervals as shown in Fig. \ref{fig:Segment}. Instead of the straightforward approach of direct synthesis from a concatenation of all input frames and events as introduced in \cite{tulyakov2021time}, we propose to generate the final intermediate frame step-by-step. The global time interval is divided into $T$ time segments.
%
%
Given an input RGB frame $I_0$ and an event sequence \{$\mathbb{E}_{0 \rightarrow t}$\}, we obtain the sliced sequences \{$\mathbb{E}_{0 \rightarrow \frac{1}{T} t}$, $\mathbb{E}_{\frac{1}{T} t \rightarrow \frac{2}{T} t},$ $\ldots$, $\mathbb{E}_{\frac{i - 1}{T} t \rightarrow \frac{i}{T} t}$, $\ldots$, $\mathbb{E}_{\frac{T-1}{T} t \rightarrow t}$\}. We first synthesize a transitional frame $\hat{I}^{\text{proxy}}_{0 \rightarrow \frac{1}{T} t}$ using $I_0$ and $\mathbb{E}_{0 \rightarrow \frac{1}{T} t}$. We then continue and repeat the above process gradually to synthesize the desired intermediate frame $\hat{I}_{0 \rightarrow t}^{\text{proxy}}$ from this direction:
\begin{equation}
    \left\{
    \begin{array}{cl}
        \hat{I}^{\text{proxy}}_{0 \rightarrow \frac{1}{T} t} &= f_2(I_0, \mathbb{E}_{0 \rightarrow \frac{1}{T} t}), \\
         & \vdots  \\
        \hat{I}^{\text{proxy}}_{0 \rightarrow \frac{i}{T} t} &= f_2(\hat{I}^{\text{proxy}}_{0 \rightarrow \frac{i - 1}{T} t}, \mathbb{E}_{\frac{i - 1}{T} t \rightarrow \frac{i}{T} t}), \\
         & \vdots  \\
        \hat{I}_{0 \rightarrow t}^{\text{proxy}} &= f_2(\hat{I}^{\text{proxy}}_{0 \rightarrow \frac{T-1}{T} t}, \mathbb{E}_{\frac{T-1}{T} t \rightarrow t}),
    \end{array}
    \right.
\end{equation}
where $f_2$ is a neural network function.
    


The reverse direction adopts a similar process to generate the intermediate frame $\hat{I}_{1 \rightarrow t}^{\text{proxy}}$. 

Event cameras provide quasi-continuous recordings of true visual signals between two RGB frames. If we only conduct direct synthesis from the complete event sequence, the quality of the interpolated frame will suffer in challenging situations in which substantial noise may have occurred. 
The proposed \emph{Transitional Synthesis Branch} enables the model to learn from local time intervals within which the motion is simpler and noise accumulations are less apparent. A different problem is hence divided into multiple similar, simpler problems. The network implicitly focuses on the quality of intermediate results to ensure that the final cumulative result is more accurate. The \emph{Transitional Synthesis Branch} takes advantage of the continuous-time nature of the event sequence to break a hard learning problem into a sequence of similar, simpler learning problems, and thus provides an interesting novel direction for event-based learning algorithms. Note that in this particular work, we use $T=2$.

\noindent \textbf{Synthesis Fusion.} After acquiring four interpolation candidates ($\hat{I}_{0 \rightarrow t}^{\text{direct}}$, $\hat{I}_{1 \rightarrow t}^{\text{direct}}$, $\hat{I}_{0 \rightarrow t}^{\text{proxy}}$ and $\hat{I}_{1 \rightarrow t}^{\text{proxy}}$) from the \emph{Direct Synthesis} and \emph{Transitional Synthesis} branches, we apply the \emph{Synthesis Fusion} module to generate the final prediction $\hat{I}_{t}^{\text{syn}}$ by fusing the results through an additional neural network (see Fig. \ref{fig:synthesis}). 
The \emph{Synthesis Fusion} block combines the results of two branches to cover both global and local time intervals. Taking the \emph{Direct Synthesis Module} into consideration is of vital importance because the information utilized within the \emph{Transitional Synthesis} branches is local and limited. The \emph{Direct Synthesis Module} learns from the complete temporal interval and helps to build connections between the isolated predictions from separate small intervals, thereby helping to ensure consistency within each sub-stage's prediction in the transitional branch. Owing to the combined design, our temporal pyramid-like architecture yields robust and consistent interpolation results.


\noindent \textbf{Loss Function.} We employ the \emph{$\ell_1$} loss as the reconstruction loss and the \emph{perceptual loss} \cite{zhang2018unreasonable} to supervise the results from both branches and the final prediction. The reconstruction loss $L_r$ is defined as
\begin{equation}
    \begin{aligned}
            L_r &= ||\hat{I}_{t}^{\text{syn}} - I_t||_1 + ||\hat{I}_{0 \rightarrow t}^{\text{direct}} - I_t||_1 + ||\hat{I}_{1 \rightarrow t}^{\text{direct}} - I_t||_1  \\
            &+ ||\hat{I}_{0 \rightarrow t}^{\text{proxy}} - I_t||_1 + ||\hat{I}_{1 \rightarrow t}^{\text{proxy}} - I_t||_1,
    \end{aligned}
\end{equation}
where $I_t$ is ground truth, and the \emph{perceptual loss} $L_{\text{perceptual}}$ follows \cite{niklaus2017video,tulyakov2021time} to constrain the similarity between prediction and ground truth from features outputted from the pretrained VGG-16 model.
Combined, the synthesis loss $L_s$ is given by
\begin{equation}
\label{lambda}%
    \begin{aligned}
        L_s &= L_r + \lambda_1 L_{\text{perceptual}},
    \end{aligned}
\end{equation}
where $\lambda_1$ is a weight factor to balance both loss functions. 

\begin{figure}[b!]
\begin{center}
\includegraphics[width=\linewidth]{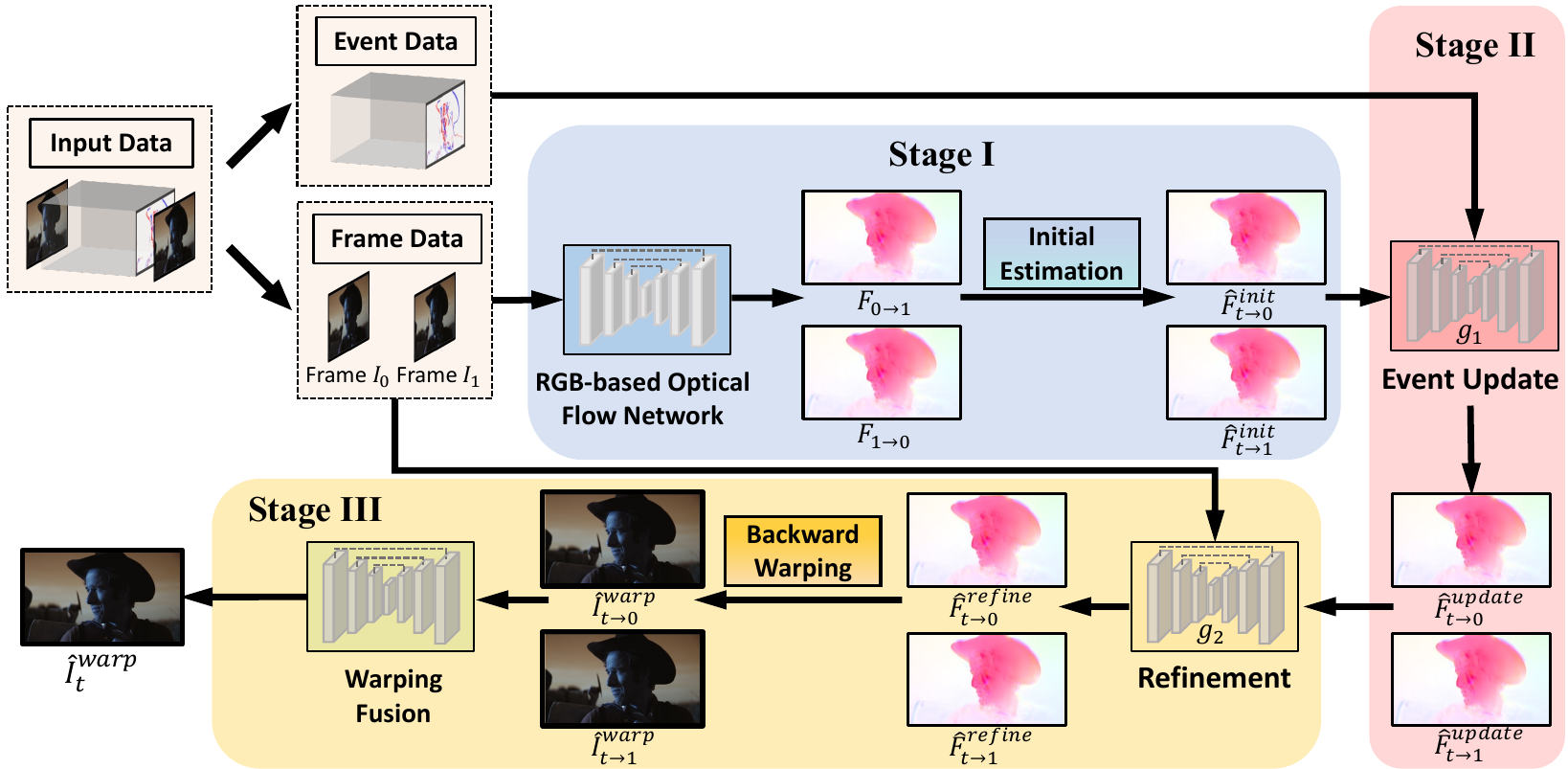}
\end{center}
\caption{Architecture of the proposed \emph{Event-guided Recurrent Warping Module}. Stage I conducts initial RGB-based optical flow estimation, and Stage II utilizes events to update the estimated optical flow. Finally, Stage III performs refinement and backward warping to obtain the interpolated frame.}
\label{fig:warping}
\end{figure}

\subsection{Event-guided Recurrent Warping Module \label{3.4}}

In light of an event camera's ability to sense densely in time and their high sensitivity with respect to image motion, we propose to utilize event sequences to guide and refine the optical flow estimated from two boundary RGB frames $I_0$ and $I_1$ instead of computing optical flow solely from event sequences (cf. Time Lens \cite{tulyakov2021time}). As demonstrated in Fig. \ref{fig:warping}, we adopt a three-stage architecture to generate a warping-based intermediate video frame. In the first stage, we initialize bi-directional optical flow through an RGB-based optical flow prediction network. Subsequently, the second stage utilizes event sequences to update the previously estimated initial optical flow. Finally, the third stage again adds the two boundary RGB frames to re-refine the estimated bi-directional optical flow. Given the refined optical flow and the two boundary frames, we then utilize backward warping and fusion to obtain the interpolated intermediate frame.

\noindent \textbf{Initial RGB-based Estimation:} Given input frames $I_0$ and $I_1$, we first employ a pre-trained RGB-based optical flow estimation network (GMFlow \cite{xu2022gmflow}) to predict bi-directional optical flow $F_{0 \rightarrow 1}$ and $F_{1 \rightarrow 0}$. Following \cite{jiang2018super}, we then approximate initial time-parametrized optical flow fields $\hat{F}_{t \rightarrow 0}^{\text{init}}$ and $\hat{F}_{t \rightarrow 1}^{\text{init}}$ through an interpolation inspired by linear fusion: 
\begin{equation}
    \begin{aligned}
        \hat{F}_{t \rightarrow 0}^{\text{init}} &= -(1-t)t{F}_{0 \rightarrow 1} + t^2F_{1 \rightarrow 0} \\
        \hat{F}_{t \rightarrow 1}^{\text{init}} &= (1-t)^2{F}_{0 \rightarrow 1} - t(1-t)F_{1 \rightarrow 0}.
    \end{aligned}
\end{equation}
Given that RGB cameras sample sparsely in time, this simple linear interpolation based temporal parametrization of the optical flow is approximate and leads to obvious limitations when modeling complicated non-linear motion. Hence, we further propose to utilize event sequences for refinement.

\noindent \textbf{Event-based Update:} Given initial optical flow results $\hat{F}_{t \rightarrow 0}^{\text{init}}$ and $\hat{F}_{t \rightarrow 1}^{\text{init}}$, we utilize events $\mathbb{E}_{t \rightarrow 0}$ and $\mathbb{E}_{1 \rightarrow t}$ (computed using the reversing strategy mentioned in Sec. \ref{3.3}) to learn event-guided residual optical flows through a neural network $g_1$:
\begin{equation}
    \begin{aligned}
        \Delta \hat{F}_{t \rightarrow 0}^{\text{event}} &= g_1(\hat{F}_{t \rightarrow 0}^{\text{init}}, \mathbb{E}_{t \rightarrow 0})\\
        \Delta \hat{F}_{t \rightarrow 1}^{\text{event}} &= g_1(\hat{F}_{t \rightarrow 1}^{\text{init}}, \mathbb{E}_{t \rightarrow 1}).
    \end{aligned}
\end{equation}
Updated optical flows are then given by:
\begin{equation}
    \begin{aligned}
        \hat{F}_{t \rightarrow 0}^{\text{update}} &= \hat{F}_{t \rightarrow 0}^{\text{init}} + \Delta \hat{F}_{t \rightarrow 0}^{\text{event}}\\
        \hat{F}_{t \rightarrow 1}^{\text{update}} &= \hat{F}_{t \rightarrow 1}^{\text{init}} + \Delta \hat{F}_{t \rightarrow 1}^{\text{event}}.
    \end{aligned}
\end{equation}
Our approach improves the use of event sequences to estimate optical flow when compared against \cite{tulyakov2021time}, which directly yields optical flow solely from events. The high sensitivity of event cameras leads to an abundance of events within short time intervals. Accumulation of substantial noise accompanied by the spatially sparse distribution of events complicates the estimation of optical flow purely based on event sequences. By recording the high-frequent brightness changes in the image, event cameras inherently serve better to sense instantaneous motion rather than to reveal pixel-level invariant correspondences based on accumulated images. We therefore propose the alternative of utilizing the rich motion information of event sequences to guide and constrain the optical flow estimated from two RGB frames.

\noindent \textbf{Refinement and Backward Warping.} The updated optical flow fields still perform poorly along motion boundaries where non-smoothness in the optical flow fields occurs \cite{jiang2018super}. Motivated by \cite{li2020video}, we concatenate flow fields $\hat{F}_{t \rightarrow 0}^{\text{update}}$ and $\hat{F}_{t \rightarrow 1}^{\text{update}}$ with boundary RGB frames $I_0$ and $I_1$ and feed them into a neural network $g_2$ to compute residual optical flows $\Delta \hat{F}_{t \rightarrow 0}$ and $\Delta \hat{F}_{t \rightarrow 1}$:
\begin{equation} 
    \begin{aligned}
        \Delta \hat{F}_{t \rightarrow 0}, \Delta \hat{F}_{t \rightarrow 1} &= g_2(\hat{F}_{t \rightarrow 0}^{\text{update}}, \hat{F}_{t \rightarrow 1}^{\text{update}}, I_0, I_1).\\
    \end{aligned}
\end{equation}
Next, we compute the re-refined optical flow field using
\begin{equation}
    \begin{aligned}
        \hat{F}_{t \rightarrow 0}^{\text{refine}}  &= \hat{F}_{t \rightarrow 0}^{\text{update}} + \Delta \hat{F}_{t \rightarrow 0} \\
        \hat{F}_{t \rightarrow 1}^{\text{refine}}  &= \hat{F}_{t \rightarrow 1}^{\text{update}} + \Delta \hat{F}_{t \rightarrow 1}.
    \end{aligned}
\end{equation}
To conclude, we backward warp \cite{jiang2018super,bao2019depth,park2020bmbc} RGB frames $I_0$ and $I_1$ using the refined optical flow fields $\hat{F}_{t \rightarrow 0}^{\text{refine}}$ and $\hat{F}_{t \rightarrow 1}^{\text{refine}}$ through differentiable interpolation \cite{jaderberg2015spatial}, respectively. This generates the two frames $\hat{I}_{t \rightarrow 0}^{\text{warp}}$ and $\hat{I}_{t \rightarrow 1}^{\text{warp}}$ at timestamp $t$, which we feed into a warp fusion network to yield the intermediate frame $\hat{I}^{\text{warp}}_t$.

\noindent \textbf{Loss Function.} The \emph{$\ell_1$} loss is employed to supervise the interpolated results both before and after the fusion module. More specifically, the warping loss $L_w$ is defined as:
\begin{equation}
    \begin{aligned}
        L_w &= ||\hat{I}^{\text{warp}}_t - I_t||_1 + ||\hat{I}_{t \rightarrow 0}^{\text{warp}} - I_t||_1 + ||\hat{I}_{t \rightarrow 1}^{\text{warp}} - I_t||_1,
    \end{aligned}
\end{equation}
where $I_t$ is ground truth.

\subsection{Attention-based Averaging Module\label{3.5}}

Our \emph{Proxy-guided Synthesis} Module is flow-free. It shows an improved ability in handling challenging situations such as fast non-linear motion, illumination changes or new object occurrences. The reason is attributed to the auxiliary visual information provided by the dynamic information in the form of events \cite{tulyakov2021time}. However, given that the synthesis module synthesizes new frames directly from events, it has defects along edges caused by noise in the events and insufficient sensitivity in low-texture regions. The \emph{Event-guided Recurrent Warping} module complements the behavior in these regions, but relies on the brightness constancy assumption, which has limited performance in the before mentioned challenging situations. We add the \emph{Attention-based Averaging} module to compensate for individual weaknesses and combine the advantages of the synthesis and warping modules \cite{tulyakov2021time}. Similar to \cite{niklaus2020softmax,jiang2018super,park2020bmbc,tulyakov2021time}, the \emph{Attention-based Averaging} module takes the output frames of the \emph{Proxy-guided Synthesis} module $\hat{I}^{\text{syn}}_t$ and the \emph{Event-guided Recurrent Warping} module $\hat{I}^{\text{warp}}_t$, and learns weights to blend the results in a pixel-wise fashion and yield the final interpolated result $\hat{I}_t$ with reduced distortions and motion blur. We adopt the same loss functions as in Sec. \ref{3.3} to supervise the final result $\hat{I}_t$.



\begin{figure*}[h!]
\vspace{0.15cm}
\begin{center}
\includegraphics[width=0.6\linewidth]{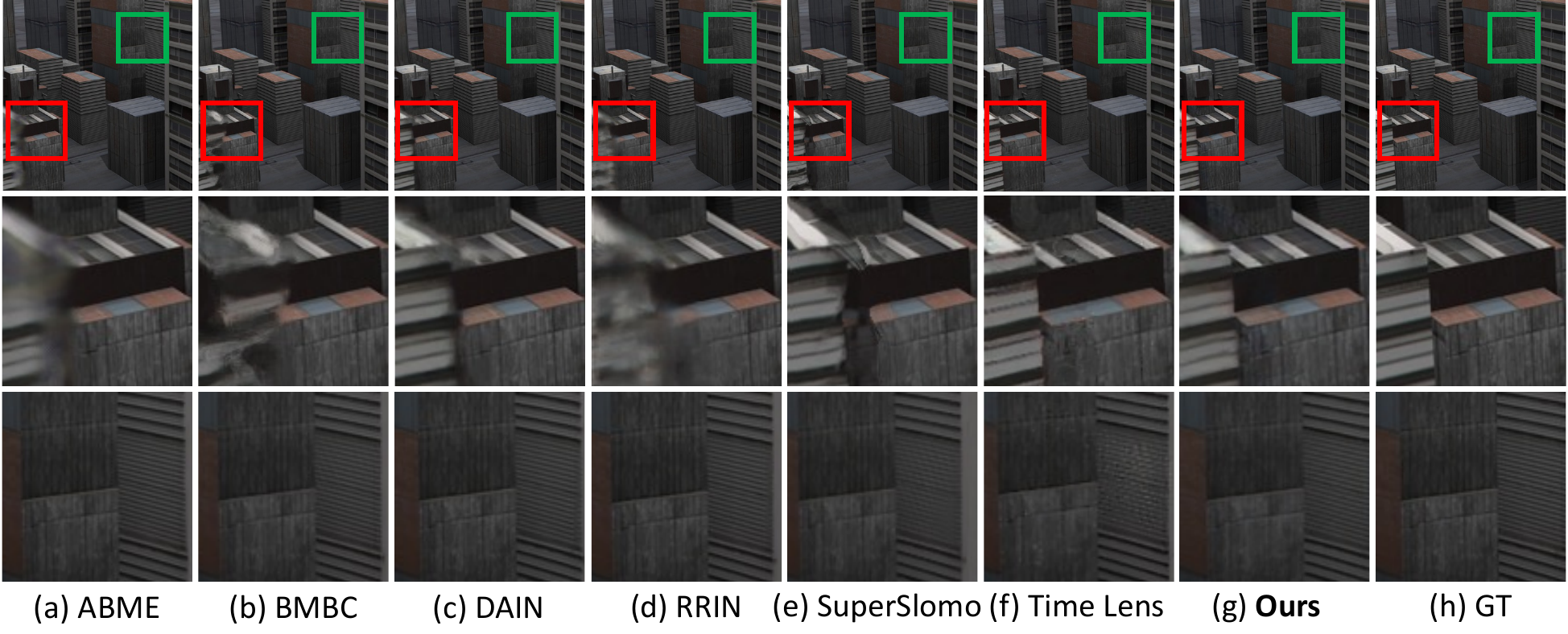}
\end{center}
\caption{Qualitative comparison on Middlebury dataset \cite{baker2011database}. Our method yields interpolated frame without blur and noticeable artifacts in the \textcolor{red}{red} boxes, and preserves detailed information like straight lines of buildings in the \textcolor{green}{green} boxes.}
\label{fig:quali}
\end{figure*} 

\section{Experiments}

\subsection{Experimental Setup}

\noindent \textbf{Datasets.} We use the Vimeo90K Septuplet dataset \cite{xue2019video} for training. Following the same setting as \cite{tulyakov2021time}, events sequences are synthetically generated using an event simulator-based \cite{rebecq2018esim} video-to-event method \cite{gehrig2020video}. The resolution of all frames and events is $448 \times 256$. We perform the quantitative evaluation on standard VFI datasets to verify our method's robustness and generalization ability. We do not conduct fine-tuning for each dataset. Specifically, we test on datasets with frames and synthetic events constructed as in \cite{tulyakov2021time}, which includes the Vimeo90K Triplet testing set \cite{xue2019video} and Middlebury \cite{baker2011database}. Moreover, again adhering to the evaluation method used in \cite{tulyakov2021time}, we perform experiments on datasets with frames and real events, including the High Quality Frames (HQF) dataset \cite{stoffregen2020reducing} and the High Speed Event-RGB (HS-ERGB) dataset \cite{tulyakov2021time}.

\noindent \textbf{Baselines.} We compare our method against state-of-the-art video frame interpolation methods, which can be divided into two categories. The frame-based category contains DAIN \cite{bao2019depth}, RRIN \cite{li2020video}, SuperSlomo \cite{jiang2018super}, BMBC \cite{park2020bmbc}, ABME \cite{park2021asymmetric} and FLAVR \cite{kalluri2020flavr}. And the event-based category contains Time Lens \cite{tulyakov2021time} and Time Lens++ \cite{tulyakov2022time}. 

\noindent \textbf{Evaluation Metrics.} We adopt signal-to-noise ratio (PSNR), structural similarity (SSIM) and LPIPS to evaluate the interpolation quality.

\noindent \textbf{Training Details.} We train the Proxy-guided Synthesis Module, Event-guided Recurrent Warping Module, and Attention-based Averaging Module for 40, 40 and 10 epochs, respectively. We train the first two modules separately and freeze their weights when training the third module. We use the Sintel checkpoint of GMFlow \cite{xu2022gmflow}, and is freezed during training. We use Adam optimizer with $\beta_1=0.9$ and $\beta_2=0.99$, and start with an initial learning rate of $1 \times 10^{-4}$. Cosine Annealing is employed to decay the learning rate. 
Regarding hyper-parameters, we empirically set $\lambda_1 = 1$ in Eq. (\ref{lambda}) in Sec. \ref{3.3}. We choose neural networks $f_1$, $f_2$ in Sec. \ref{3.3} and $g_1$, $g_2$ in Sec. \ref{3.4} as skip-connected Hourglass U-Net architectures \cite{jiang2018super,tulyakov2021time}.


\begin{table*}[ht!]
\setlength{\belowcaptionskip}{-0.5cm}
\scriptsize
\centering
\setlength{\tabcolsep}{1.73mm}
\begin{tabular}{lccccrcccccccc}
\hline
           & \multicolumn{7}{c}{(a) Synthetic Dataset}                                 & \multicolumn{1}{l}{} & \multicolumn{5}{c}{(b) Real-world Dataset}                \\  \cline{2-8} \cline{10-14} 
           & Triplet \cite{xue2019video}     &  & \multicolumn{2}{c}{Septuplet \cite{xue2019video}} &  & \multicolumn{2}{c}{Middlebury \cite{baker2011database}} &                      & \multicolumn{2}{c}{HQF \cite{stoffregen2020reducing}}   &  & \multicolumn{2}{c}{HS-ERGB (close) \cite{tulyakov2021time}} \\ \cline{2-2} \cline{4-5} \cline{7-8} \cline{10-11} \cline{13-14} 
           & x2          &  & x2            & x4            &  & x2             & x4            &                      & x2          & x4          &  & x6            & x8            \\ \cline{1-14}  
SuperSlomo \cite{jiang2018super} & 33.44/0.951 &  & 33.96/0.943   & 29.44/0.888   &  & 29.68/0.876    & 26.42/0.819   &                      & 28.76/0.861 & 25.54/0.761 &  & 28.35/0.788   & 27.27/0.755   \\
RRIN \cite{li2020video}       & 34.68/0.962 &  & 35.56/0.954   & 29.41/0.891   &  & 31.17/0.894    & 27.28/0.841   &                      & 29.76/0.874 & 26.11/0.778 &  & 28.70/0.813   & 27.44/0.800   \\
BMBC \cite{park2020bmbc}       & 35.09/0.963 &  & 35.48/0.949   & 30.33/0.897   &  & 30.71/0.889    & 26.45/0.821   &                      & \textcolor{blue}{30.74}/0.875 & 27.01/0.781 &  & 29.32/0.821   & 27.89/0.808   \\
ABME \cite{park2021asymmetric}       & 36.22/\textcolor{red}{0.969} &  & 36.53/0.955   & -             &  & 31.66/0.900    & -             &                      & 30.58/0.880 & -           &  & -   & -             \\
DAIN \cite{bao2019depth}       & 34.70/0.964 &  & 35.29/0.954   & 29.87/0.900   &  & 30.90/0.896    & 26.65/0.831   &                      & 29.82/0.875 & 26.10/0.782 &  & 29.03/0.807   & 28.50/0.801   \\
Time Lens \cite{tulyakov2021time}   & \textcolor{blue}{36.31}/0.962 &  & \textcolor{blue}{36.87}/\textcolor{blue}{0.960}   & \textcolor{blue}{35.58}/\textcolor{blue}{0.949}   &  & \textcolor{red}{33.27}/\textcolor{red}{0.929}    & \textcolor{red}{32.13}/\textcolor{red}{0.908}   &                      & 30.57/\textcolor{blue}{0.903} & \textcolor{red}{28.98}/\textcolor{red}{0.873} &  & \textcolor{blue}{32.19}/\textcolor{blue}{0.839}   & \textcolor{blue}{31.68}/\textcolor{blue}{0.835}    \\
\textbf{Ours}       & \textcolor{red}{36.56}/\textcolor{blue}{0.965} &  & \textcolor{red}{38.14}/\textcolor{red}{0.968}   & \textcolor{red}{36.34}/\textcolor{red}{0.960}   &  & \textcolor{blue}{32.51}/\textcolor{blue}{0.909}    & \textcolor{blue}{31.01}/\textcolor{blue}{0.886}   &                      & \textcolor{red}{31.75}/\textcolor{red}{0.910} & \textcolor{blue}{28.56}/\textcolor{blue}{0.850} &  & \textcolor{red}{33.21}/\textcolor{red}{0.847}   & \textcolor{red}{32.95}/\textcolor{red}{0.844}   \\ \hline
\end{tabular}

\caption{Quantitative results on (a) synthetic datasets: Vimeo 90k Triplet \cite{xue2019video}, Vimeo 90k Setuplet \cite{xue2019video} and Middlebury \cite{baker2011database}; (b) real datasets: HQF dataset \cite{stoffregen2020reducing}, and the close sequences of the HS-ERGB dataset \cite{tulyakov2021time} ($\times 2$, $\times 4$, $\times 6$, $\times 8$ denote skip 1, 3, 5, 7 frames, respectively). \textcolor{red}{Red} and \textcolor{blue}{blue} indicate the best and second best performance, respectively.}
\label{tab:quanti}%
\end{table*}

\begin{table}
\setlength{\belowcaptionskip}{-0.7cm}
\tiny
        \centering
        \renewcommand{\tabcolsep}{2pt}
        \begin{tabular}{c|cc}
        \toprule
        \textbf{Method}        & \textbf{PSNR$\uparrow$} & \textbf{LPIPS$\downarrow$} \\ 
        \midrule
        FLAVR \cite{kalluri2020flavr} & 27.42 &0.031 \\
        SuperSlomo \cite{jiang2018super}      & 30.05         & 0.103        \\
        Timelens \cite{tulyakov2021time}      & 33.48       & 0.017         \\ 
        Timelens++ \cite{tulyakov2022time} &33.09 &0.016 \\
        \midrule
        \textbf{Ours} &\textbf{33.53} &\textbf{0.015} \\
        \bottomrule
        \end{tabular}
        \caption{ Quantitative results on HS-ERGB dataset \cite{tulyakov2021time}. We conduct $\times$8 interpolation, and average the score over close and far subset.}
        \label{ablat:hsergb}
\end{table}

\subsection{Comparisons Against State-of-the-art Methods}

\noindent \textbf{Evaluations on datasets with synthetic events.}
Table \ref{tab:quanti} shows that our method outperforms both frame-based and event-based state-of-the-art algorithms on the Vimeo90K Septuplet dataset \cite{xue2019video} in terms of both single-frame interpolation and multi-frame interpolation by a significant margin. 
For frame-based methods, our approach exceeds SuperSlomo\cite{jiang2018super} and RRIN\cite{li2020video} which approximate bi-directional optical flow, and ABME \cite{park2021asymmetric} which estimates bilateral motion fields. This demonstrates the advantage of the auxiliary visual information introduced by the highly dynamic event cameras. Our model also outperforms the event-based method Time Lens \cite{tulyakov2021time}, which justifies the superiority the design choices we made in our synthesis and warping modules. To verify the generalization ability of our method, we also evaluate our method on Middlebury \cite{baker2011database} and Vimeo90K (triplet) \cite{xue2019video}, where our method also demonstrates favorable performance.
Fig. \ref{fig:quali} illustrates qualitative comparisons on the Middlebury dataset \cite{baker2011database}. As can be observed, our method maintains good performance while baselines produce blurry results or noticeable artifacts.\\
\noindent \textbf{Evaluations on datasets with real events.}
 To evaluate the adaptability of our method to real events, we assess performance of both single and multi-frame interpolation on the HQF \cite{stoffregen2020reducing} and close sequences of HS-ERGB datasets \cite{tulyakov2021time}. Results are summarized in Table \ref{tab:quanti}. The results on datasets with real events are consistent with the results obtain for synthetic events, although the synthetic-to-real gap necessarily causes a slight drop in overall performance with respect to the results we obtained on synthetic datasets. Furthermore, we compare with recently proposed Timelens++ \cite{tulyakov2022time} on HS-ERGB dataset \cite{tulyakov2021time} in Table \ref{ablat:hsergb}, where we conduct $\times8$ interpolation and average the score over close and far subsets. The results are consistent with those obtained for single-frame interpolation, and our method keeps outperforming the state-of-the-art. Finally, we captured demo videos of daily scenes with complex motion using a DAVIS 346 event camera, please refer to the supplemental video.
 
\begin{table*}[htbp]
\tiny
\vspace{0.25cm}
    \centering
    \begin{subtable}{0.19\linewidth}
        \centering
        \begin{tabular}{c|cc}
        \toprule
        \textbf{Module}                & \textbf{PSNR$\uparrow$} & \textbf{SSIM$\uparrow$} \\ \hline
        Warping      & 35.77         & 0.964         \\
        Synthesis & 37.71         & 0.959         \\
        Averaging      & 38.14         & 0.968         \\ \hline
        \end{tabular}
        \caption{}
        \label{ablat:a}
    \end{subtable}
    \begin{subtable}{0.19\linewidth}
        \centering
        \begin{tabular}{c|cc}
        \toprule
        \textbf{\# Proxies}        & \textbf{PSNR$\uparrow$} & \textbf{SSIM$\uparrow$} \\ \hline
        1       & 35.87         & 0.954         \\
        2  & 35.95         & 0.959         \\
        4        & 34.61         & 0.943         \\ \hline
        \end{tabular}
        \caption{}
        \label{ablat:b}
    \end{subtable}
    \begin{subtable}{0.19\linewidth}
        \centering
        \begin{tabular}{c|cc}
        \toprule
        \textbf{Case}        & \textbf{PSNR$\uparrow$} & \textbf{SSIM$\uparrow$} \\ \hline
        only 1      & 35.87         & 0.954         \\
        1 \& 2     & 37.71         & 0.959         \\
        1 \&2 \& 4       & 37.78         & 0.956         \\ \hline
        \end{tabular}
        \caption{}
        \label{ablat:c}
    \end{subtable}
    \begin{subtable}{0.19\linewidth}
        \centering
        \begin{tabular}{c|cc}
        \toprule
        \textbf{Case}        & \textbf{PSNR$\uparrow$} & \textbf{SSIM$\uparrow$} \\ \hline
        RGB-guided      & 33.74         & 0.938         \\
        event-guided       & 35.77         & 0.964        \\ \hline
        \end{tabular}
        \caption{}
        \label{ablat:d}
    \end{subtable}
    \begin{subtable}{0.19\linewidth}
        \centering
        \begin{tabular}{c|cc}
        \toprule
        \textbf{Case}        & \textbf{PSNR$\uparrow$} & \textbf{SSIM$\uparrow$} \\ \hline
        w/o event      & 32.24         & 0.941         \\
        w event       & 35.77         & 0.964         \\ \hline
        \end{tabular}
        \caption{}
        \label{ablat:e}
    \end{subtable}
    \caption{\textbf{Ablation results.} Ablation studies of our method to evaluate (a) the performance of each module, (b) different event segmentation strategies (division into 1,2, or 4 proxies, respectively), (c) different fusion strategies, (d) RGB-guided warping vs event-guided warping , (e) the importance of event-based updating.}
    \label{tab:ablation}
\end{table*}
\begin{figure}[h]
\setlength{\belowcaptionskip}{-0.5cm}
\begin{center}
\includegraphics[width=\linewidth]{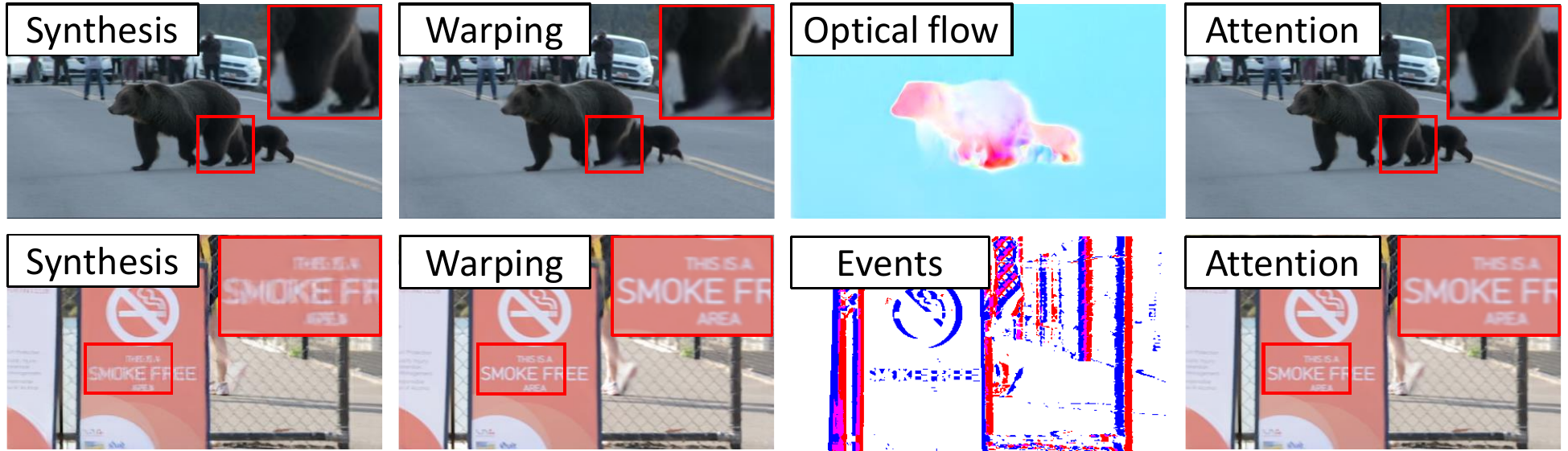}
\end{center}
\vspace{-0.3cm}
\caption{Visualization of interpolation results from the different modules (\emph{best viewed when zoomed-in)}.}
\label{fig:component}
\end{figure}
\begin{figure}[htbp]
\setlength{\belowcaptionskip}{-0.5cm}
\begin{center}
\includegraphics[width=0.7\linewidth]{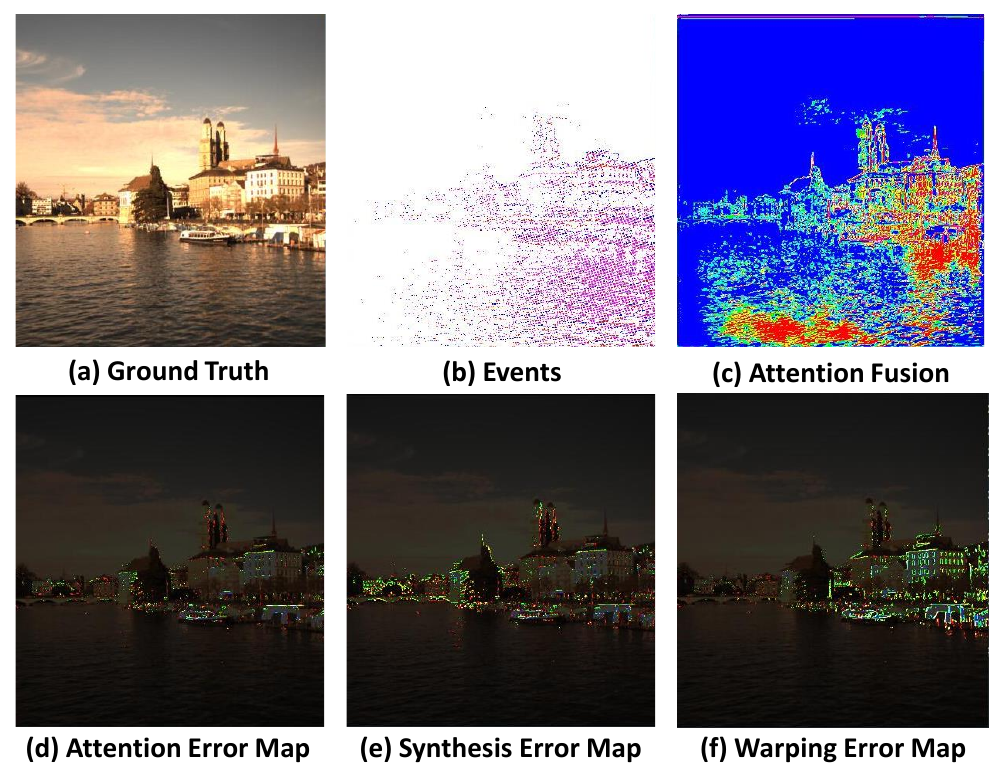}
\end{center}
\caption{Visualizations of interpolation results from the different modules on the HS-ERGB dataset \cite{tulyakov2021time}. The corresponding events, attention fusion map and interpolation eror maps for each module are also illustrated.}
\label{fig:zurich}
\end{figure}
 \subsection{Ablation Experiments}
We present a detailed ablation study of our method in Table. \ref{tab:ablation}. We analyze the contributions of the three key modules to the final interpolation and compare several architecture design choices to assess their effects. Ablations are conducted on the Vimeo90K Septuplet dataset \cite{xue2019video}.\\
\noindent \textbf{The impact of the different modules.} To analyze the contribution of our three modules, we show the interpolation quality of each module in Table \ref{ablat:a}. The result demonstrates that both our synthesis and warping module achieve good interpolation results. Besides, it shows that our attention-based averaging module successfully combines the advantages of the two previous components, resulting in an increase in PSNR and SSIM to 38.14 and 0.968, respectively. Fig. \ref{fig:component} visualizes the impacts of the different modules. The first row of Fig. \ref{fig:component} shows an example where our warping module yields blurry results due to insufficient ability to model overlapping motion. In contrast, our attention module predicts correct results using the better results of the synthesis module. The second row of Fig. \ref{fig:component} shows an example where our synthesis module generates blurry results due to event noise. However, the attention module again interpolates clear results by giving preference to the good predictions of the warping module. Furthermore, we present visualization results on the HS-ERGB dataset \cite{tulyakov2021time} in Fig. \ref{fig:zurich}. The interpolation error maps demonstrate that our attention-based averaging module effectively leverages the synthesis module's better results for nearby objects and the warping module's superior results for distant objects. Consequently, our approach achieves significantly fewer errors in the final output.\\

\noindent \textbf{Ablations of the Proxy-guided Synthesis Module}

    \noindent \textbf{\textit{The impact of different event segmentation strategies.}} 
    Table. \ref{ablat:b} reports results for different numbers of proxies dividing the original event sequence. As shown, we obtain best performance when the number of proxies is set to 2, followed by 1 and then 4. When the number of proxies is set to 1, we directly feed the whole event sequence into the neural network without any segmentation. The accumulated grid-based representation of the events results in a loss of the quasi-continuous, high-frequent temporal information provided by the original events, thus resulting in worse performance. When setting the number of proxies to four ($T=4$), the performance drops as the event representation becomes very sparse, thereby providing insufficient information for detailed frame synthesis. The ideal proxy number remains a function of the original video framerate. 
    
    \noindent \textbf{\textit{The impact of different fusion strategies.}} Table. \ref{ablat:c} shows results of different fusion choices in the synthesis module. As discussed in Sec. \ref{3.3}, we propose to segment events into $T$ proxies. We compare the choices of: 1) only $T$ = 1, 2) fuse $T$ = 1 and $T$ = 2 from two individual branches, and 3) fuse $T$ = 1, $T$ = 2 and $T$ = 4 from three individual branches. Table \ref{ablat:c} shows that another improvement can be achieved by combining an additional branch of $T$ = 4, which however comes at an unjustifiable increase in computational cost during both training and testing.\\

\noindent \textbf{Ablations of the Event-guided Warping Module}\\
\noindent \textbf{\textit{The impact of event-guided optical flow.}} To justify the core design of our \emph{Event-guided Recurrent Warping Module}, we check the alternative of using images to refine a purely event-based initial flow, and show the results in Table \ref{ablat:d}. 
    We first replace our image-based optical flow network with the pre-trained warping network of \cite{tulyakov2021time} (trained on the same dataset as ours), which estimates optical flow solely from events. 
    We then use images to refine the estimated flow instead of the original \emph{Event Update} block. For the sake of a fair comparison, this alternative is trained for the same number of epochs than the original network. As shown in Table. \ref{ablat:d}, our method performs better than the image-guided event-based flow estimation, which confirms our design choice. 
    \noindent \textbf{\textit{The importance of the Event Update block.}} We evaluate the effectiveness of the \emph{Event Update} block in Sec. \ref{3.4} by training a warping-based module without event information. It directly uses the pre-trained GMFlow optical flow network \cite{xu2022gmflow} and skips the \emph{Event Update} block, while leaving the remaining architecture unchanged. Table. \ref{ablat:e} shows the results obtained by this variant, which are much worse than the original ones (3.53 dB drop in terms of PSNR). This confirms the importance of the \emph{Event Update} block.

\section{Conclusion}
The present work shows that a careful consideration of the inherent properties of event cameras in the design of neural architectures can help to improve prediction results, in this case for video frame interpolation. Event cameras only encode changes of intensity and polarity rather than absolute color intensities, and thus prove to be a better choice for RGB based optical flow prediction refinement rather than for direct optical flow prediction. Furthermore, in light of the quasi-continuous nature of event streams, we propose an incremental synthesis strategy that breaks down the global prediction into multiple simpler and equivalent short-term prediction steps. Extensive experiments on both synthetic and real-world datasets show that our method demonstrates favorable VFI performance. We believe that the ideas communicated in this work can be replicated towards alternative event-based tasks, thereby making a strong contribution towards the combined use of regular and dynamic vision sensors.

{\small
\bibliographystyle{IEEEtran}
\bibliography{egbib}

\begin{thebibliography}{10}
\providecommand{\url}[1]{#1}
\csname url@rmstyle\endcsname
\providecommand{\newblock}{\relax}
\providecommand{\bibinfo}[2]{#2}
\providecommand\BIBentrySTDinterwordspacing{\spaceskip=0pt\relax}
\providecommand\BIBentryALTinterwordstretchfactor{4}
\providecommand\BIBentryALTinterwordspacing{\spaceskip=\fontdimen2\font plus
\BIBentryALTinterwordstretchfactor\fontdimen3\font minus
  \fontdimen4\font\relax}
\providecommand\BIBforeignlanguage[2]{{%
\expandafter\ifx\csname l@#1\endcsname\relax
\typeout{** WARNING: IEEEtran.bst: No hyphenation pattern has been}%
\typeout{** loaded for the language `#1'. Using the pattern for}%
\typeout{** the default language instead.}%
\else
\language=\csname l@#1\endcsname
\fi
#2}}

\bibitem{niklaus2017video}
S.~Niklaus, L.~Mai, and F.~Liu, ``Video frame interpolation via adaptive
  convolution,'' in \emph{Proceedings of the IEEE Conference on Computer Vision
  and Pattern Recognition}, 2017, pp. 670--679.

\bibitem{niklaus2018context}
S.~Niklaus and F.~Liu, ``Context-aware synthesis for video frame
  interpolation,'' in \emph{Proceedings of the IEEE conference on computer
  vision and pattern recognition}, 2018, pp. 1701--1710.

\bibitem{lin2020learning}
S.~Lin, J.~Zhang, J.~Pan, Z.~Jiang, D.~Zou, Y.~Wang, J.~Chen, and J.~Ren,
  ``Learning event-driven video deblurring and interpolation,'' in
  \emph{European Conference on Computer Vision}.\hskip 1em plus 0.5em minus
  0.4em\relax Springer, 2020, pp. 695--710.

\bibitem{han2021evintsr}
J.~Han, Y.~Yang, C.~Zhou, C.~Xu, and B.~Shi, ``Evintsr-net: Event guided
  multiple latent frames reconstruction and super-resolution,'' in
  \emph{Proceedings of the IEEE/CVF International Conference on Computer
  Vision}, 2021, pp. 4882--4891.

\bibitem{yu2021training}
Z.~Yu, Y.~Zhang, D.~Liu, D.~Zou, X.~Chen, Y.~Liu, and J.~S. Ren, ``Training
  weakly supervised video frame interpolation with events,'' in
  \emph{Proceedings of the IEEE/CVF International Conference on Computer
  Vision}, 2021, pp. 14\,589--14\,598.

\bibitem{tulyakov2021time}
S.~Tulyakov, D.~Gehrig, S.~Georgoulis, J.~Erbach, M.~Gehrig, Y.~Li, and
  D.~Scaramuzza, ``Time lens: Event-based video frame interpolation,'' in
  \emph{Proceedings of the IEEE/CVF Conference on Computer Vision and Pattern
  Recognition}, 2021, pp. 16\,155--16\,164.

\bibitem{gallego2020event}
G.~Gallego, T.~Delbr{\"u}ck, G.~Orchard, C.~Bartolozzi, B.~Taba, A.~Censi,
  S.~Leutenegger, A.~J. Davison, J.~Conradt, K.~Daniilidis, \emph{et~al.},
  ``Event-based vision: A survey,'' \emph{IEEE transactions on pattern analysis
  and machine intelligence}, vol.~44, no.~1, pp. 154--180, 2020.

\bibitem{lichtsteiner2008128}
P.~Lichtsteiner, C.~Posch, and T.~Delbruck, ``A 128 $\times$128 120 db 15$\mu$s
  latency asynchronous temporal contrast vision sensor,'' \emph{IEEE journal of
  solid-state circuits}, vol.~43, no.~2, pp. 566--576, 2008.

\bibitem{feichtenhofer2019slowfast}
C.~Feichtenhofer, H.~Fan, J.~Malik, and K.~He, ``Slowfast networks for video
  recognition,'' in \emph{Proceedings of the IEEE/CVF international conference
  on computer vision}, 2019, pp. 6202--6211.

\bibitem{jaderberg2015spatial}
M.~Jaderberg, K.~Simonyan, A.~Zisserman, \emph{et~al.}, ``Spatial transformer
  networks,'' \emph{Advances in neural information processing systems},
  vol.~28, 2015.

\bibitem{jiang2018super}
H.~Jiang, D.~Sun, V.~Jampani, M.-H. Yang, E.~Learned-Miller, and J.~Kautz,
  ``Super slomo: High quality estimation of multiple intermediate frames for
  video interpolation,'' in \emph{Proceedings of the IEEE conference on
  computer vision and pattern recognition}, 2018, pp. 9000--9008.

\bibitem{li2020video}
H.~Li, Y.~Yuan, and Q.~Wang, ``Video frame interpolation via residue
  refinement,'' in \emph{ICASSP 2020-2020 IEEE International Conference on
  Acoustics, Speech and Signal Processing (ICASSP)}.\hskip 1em plus 0.5em minus
  0.4em\relax IEEE, 2020, pp. 2613--2617.

\bibitem{xu2019quadratic}
X.~Xu, L.~Siyao, W.~Sun, Q.~Yin, and M.-H. Yang, ``Quadratic video
  interpolation,'' \emph{Advances in Neural Information Processing Systems},
  vol.~32, 2019.

\bibitem{chi2020all}
Z.~Chi, R.~Mohammadi~Nasiri, Z.~Liu, J.~Lu, J.~Tang, and K.~N. Plataniotis,
  ``All at once: Temporally adaptive multi-frame interpolation with advanced
  motion modeling,'' in \emph{European Conference on Computer Vision}.\hskip
  1em plus 0.5em minus 0.4em\relax Springer, 2020, pp. 107--123.

\bibitem{niklaus2020softmax}
S.~Niklaus and F.~Liu, ``Softmax splatting for video frame interpolation,'' in
  \emph{Proceedings of the IEEE/CVF Conference on Computer Vision and Pattern
  Recognition}, 2020, pp. 5437--5446.

\bibitem{park2020bmbc}
J.~Park, K.~Ko, C.~Lee, and C.-S. Kim, ``Bmbc: Bilateral motion estimation with
  bilateral cost volume for video interpolation,'' in \emph{European Conference
  on Computer Vision}.\hskip 1em plus 0.5em minus 0.4em\relax Springer, 2020,
  pp. 109--125.

\bibitem{park2021asymmetric}
J.~Park, C.~Lee, and C.-S. Kim, ``Asymmetric bilateral motion estimation for
  video frame interpolation,'' in \emph{Proceedings of the IEEE/CVF
  International Conference on Computer Vision}, 2021, pp. 14\,539--14\,548.

\bibitem{lu2022video}
L.~Lu, R.~Wu, H.~Lin, J.~Lu, and J.~Jia, ``Video frame interpolation with
  transformer,'' in \emph{Proceedings of the IEEE/CVF Conference on Computer
  Vision and Pattern Recognition}, 2022, pp. 3532--3542.

\bibitem{huang2022real}
Z.~Huang, T.~Zhang, W.~Heng, B.~Shi, and S.~Zhou, ``Real-time intermediate flow
  estimation for video frame interpolation,'' in \emph{European Conference on
  Computer Vision}.\hskip 1em plus 0.5em minus 0.4em\relax Springer, 2022, pp.
  624--642.

\bibitem{kalluri2020flavr}
T.~Kalluri, D.~Pathak, M.~Chandraker, and D.~Tran, ``Flavr: Flow-agnostic video
  representations for fast frame interpolation,'' \emph{arXiv preprint
  arXiv:2012.08512}, 2020.

\bibitem{bao2019depth}
W.~Bao, W.-S. Lai, C.~Ma, X.~Zhang, Z.~Gao, and M.-H. Yang, ``Depth-aware video
  frame interpolation,'' in \emph{Proceedings of the IEEE/CVF Conference on
  Computer Vision and Pattern Recognition}, 2019, pp. 3703--3712.

\bibitem{paredes2021back}
F.~Paredes-Vall{\'e}s and G.~C. de~Croon, ``Back to event basics:
  Self-supervised learning of image reconstruction for event cameras via
  photometric constancy,'' in \emph{Proceedings of the IEEE/CVF Conference on
  Computer Vision and Pattern Recognition}, 2021, pp. 3446--3455.

\bibitem{rebecq2019events}
H.~Rebecq, R.~Ranftl, V.~Koltun, and D.~Scaramuzza, ``Events-to-video: Bringing
  modern computer vision to event cameras,'' in \emph{Proceedings of the
  IEEE/CVF Conference on Computer Vision and Pattern Recognition}, 2019, pp.
  3857--3866.

\bibitem{wang2020event}
B.~Wang, J.~He, L.~Yu, G.-S. Xia, and W.~Yang, ``Event enhanced high-quality
  image recovery,'' in \emph{European Conference on Computer Vision}.\hskip 1em
  plus 0.5em minus 0.4em\relax Springer, 2020, pp. 155--171.

\bibitem{pan2019bringing}
L.~Pan, C.~Scheerlinck, X.~Yu, R.~Hartley, M.~Liu, and Y.~Dai, ``Bringing a
  blurry frame alive at high frame-rate with an event camera,'' in
  \emph{Proceedings of the IEEE/CVF Conference on Computer Vision and Pattern
  Recognition}, 2019, pp. 6820--6829.

\bibitem{zhang2022unifying}
X.~Zhang and L.~Yu, ``Unifying motion deblurring and frame interpolation with
  events,'' in \emph{Proceedings of the IEEE/CVF Conference on Computer Vision
  and Pattern Recognition}, 2022, pp. 17\,765--17\,774.

\bibitem{tulyakov2022time}
S.~Tulyakov, A.~Bochicchio, D.~Gehrig, S.~Georgoulis, Y.~Li, and D.~Scaramuzza,
  ``Time lens++: Event-based frame interpolation with parametric non-linear
  flow and multi-scale fusion,'' in \emph{Proceedings of the IEEE/CVF
  Conference on Computer Vision and Pattern Recognition}, 2022, pp.
  17\,755--17\,764.

\bibitem{wu2022video}
S.~Wu, K.~You, W.~He, C.~Yang, Y.~Tian, Y.~Wang, Z.~Zhang, and J.~Liao, ``Video
  interpolation by event-driven anisotropic adjustment of optical flow,''
  \emph{arXiv preprint arXiv:2208.09127}, 2022.

\bibitem{he2022timereplayer}
W.~He, K.~You, Z.~Qiao, X.~Jia, Z.~Zhang, W.~Wang, H.~Lu, Y.~Wang, and J.~Liao,
  ``Timereplayer: Unlocking the potential of event cameras for video
  interpolation,'' in \emph{Proceedings of the IEEE/CVF Conference on Computer
  Vision and Pattern Recognition}, 2022, pp. 17\,804--17\,813.

\bibitem{zhu2019unsupervised}
A.~Z. Zhu, L.~Yuan, K.~Chaney, and K.~Daniilidis, ``Unsupervised event-based
  learning of optical flow, depth, and egomotion,'' in \emph{Proceedings of the
  IEEE/CVF Conference on Computer Vision and Pattern Recognition}, 2019, pp.
  989--997.

\bibitem{zhang2018unreasonable}
R.~Zhang, P.~Isola, A.~A. Efros, E.~Shechtman, and O.~Wang, ``The unreasonable
  effectiveness of deep features as a perceptual metric,'' in \emph{Proceedings
  of the IEEE conference on computer vision and pattern recognition}, 2018, pp.
  586--595.

\bibitem{xu2022gmflow}
H.~Xu, J.~Zhang, J.~Cai, H.~Rezatofighi, and D.~Tao, ``Gmflow: Learning optical
  flow via global matching,'' in \emph{Proceedings of the IEEE/CVF conference
  on computer vision and pattern recognition}, 2022, pp. 8121--8130.

\bibitem{baker2011database}
S.~Baker, D.~Scharstein, J.~Lewis, S.~Roth, M.~J. Black, and R.~Szeliski, ``A
  database and evaluation methodology for optical flow,'' \emph{International
  journal of computer vision}, vol.~92, no.~1, pp. 1--31, 2011.

\bibitem{xue2019video}
T.~Xue, B.~Chen, J.~Wu, D.~Wei, and W.~T. Freeman, ``Video enhancement with
  task-oriented flow,'' \emph{International Journal of Computer Vision}, vol.
  127, no.~8, pp. 1106--1125, 2019.

\bibitem{rebecq2018esim}
H.~Rebecq, D.~Gehrig, and D.~Scaramuzza, ``Esim: an open event camera
  simulator,'' in \emph{Conference on Robot Learning}.\hskip 1em plus 0.5em
  minus 0.4em\relax PMLR, 2018, pp. 969--982.

\bibitem{gehrig2020video}
D.~Gehrig, M.~Gehrig, J.~Hidalgo-Carri{\'o}, and D.~Scaramuzza, ``Video to
  events: Recycling video datasets for event cameras,'' in \emph{Proceedings of
  the IEEE/CVF Conference on Computer Vision and Pattern Recognition}, 2020,
  pp. 3586--3595.

\bibitem{stoffregen2020reducing}
T.~Stoffregen, C.~Scheerlinck, D.~Scaramuzza, T.~Drummond, N.~Barnes,
  L.~Kleeman, and R.~Mahony, ``Reducing the sim-to-real gap for event
  cameras,'' in \emph{European Conference on Computer Vision}.\hskip 1em plus
  0.5em minus 0.4em\relax Springer, 2020, pp. 534--549.

\end{thebibliography}
}







\end{document}